\documentclass{article}
\usepackage{spconf,amsmath,graphicx}

\usepackage{mwe} 

\usepackage{graphicx}
\usepackage{textcomp}
\usepackage{amsmath}
\usepackage{breqn}
\usepackage[table]{xcolor}
\usepackage{xspace}
\usepackage{wrapfig}

\usepackage{mwe}
\usepackage{subfig}
\usepackage{float} 
\usepackage{units}
\usepackage{adjustbox}
\usepackage{rotating}

\usepackage{multirow}
\usepackage{hyperref}

\DeclareMathOperator*{\argmin}{arg\,min}
\newcommand{\MATLAB}{\textsc{Matlab}\xspace}

\title{Localization of Cochlear Implant Electrodes from Cone Beam Computed Tomography using Particle Belief Propagation}
\name{Hendrik Hachmann$^{\star}$\sthanks{H.H. and B.R. were partly funded by the Lower Saxony Ministry of Science and Culture under grant number ZN 3491 within the Lower Saxony “Vorab“ of the Volkswagen Foundation and by the Federal Ministry of Education and Research (BMBF), Germany under the project Leibniz\-KILabor (grant no. 01DD20003) and the Center for Digital Innovations (ZDIN). W.N. and B.K. received funding for this research from the Deutsche Forschungsgemeinschaft (DFG, German Research Foundation) under Germany's Excellence Strategy EXC 2177/1 'Hearing4all', DFG project numbers 390895286 and 396932747 (PI: W.N.).}
\qquad Benjamin Kr\"uger$^{\dagger}$
\qquad Bodo Rosenhahn$^{\star}$
\qquad Waldo Nogueira$^{\dagger}$}

\address{$^{\star}$ Leibniz University Hanover, Germany\\
    $^{\dagger}$ Department of Otorhinolaryngology, Hannover Medical School, Hanover, Germany\\
    $^{\dagger}$Cluster of Excellence ‘Hearing4All’, Hanover, Germany} 

\begin{document}
%
\maketitle
\begin{abstract}
Cochlear implants (CIs) are implantable medical devices that can restore the hearing sense of people suffering from profound hearing loss. The CI uses a set of electrode contacts placed inside the cochlea to stimulate the auditory nerve with current pulses. The exact location of these electrodes may be an important parameter to improve and predict the performance with these devices. Currently the methods used in clinics to characterize the geometry of the cochlea as well as to estimate the electrode positions are manual, error-prone and time consuming. 

We propose a Markov random field (MRF) model for CI electrode localization for cone beam computed tomography (CBCT) data-sets. Intensity and shape of electrodes are included as prior knowledge as well as distance and angles between contacts. MRF inference is based on slice sampling particle belief propagation and guided by several heuristics. A stochastic search finds the best maximum a posteriori estimation among sampled MRF realizations. 

We evaluate our algorithm on synthetic and real CBCT data-sets and compare its performance with two state of the art algorithms. An increase of localization precision up to 31.5\% (mean), or 48.6\% (median) respectively, on real CBCT data-sets is shown.
\end{abstract}
\section{Introduction}
Cochlear implants (CIs) are implantable medical devices that are used to restore the sense of hearing for people with profound deafness given that the auditory anatomy is fully
\begin{figure}[!htbp]
    \centering
    \includegraphics[height=0.125\textwidth]{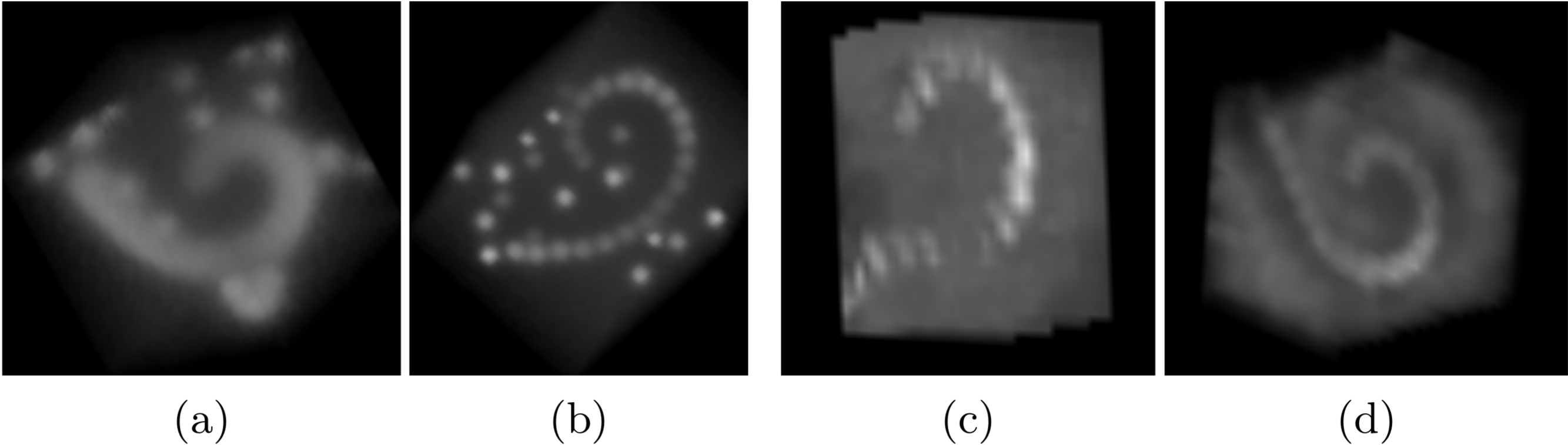}
    \caption{Comparison of 2 synthetic (a,b) and 2 real (c,d) CBCT data-sets of cochlear implants (CIs). (a) is an example of a closely-spaced EA, while (b,c,d) represents distantly-spaced EAs. Both synthetic data-sets include a high number of false positives to evaluate the robustness of localization algorithms.}
    \label{CT_datasets}
\end{figure}developed \cite{Wilson:1991}. 
A CI consists of an external sound processor and an electrode array (EA). Latter is implanted inside the cochlea, which is a spiral-shaped bone cavity making 2.5 turns around its axis with a length that ranges between 32 and \unit[43.5]{mm} \cite{WURFEL:2014}. 
The EAs are manufactured with lengths ranging from 10 to \unit[31]{mm} including 12-22 electrode contacts.
Long EAs may achieve up to two full turns of insertion to reach low-frequency stimulation \cite{Hochmair:2014}.
After surgery, CBCT scans are recorded to determine the position of the EA contacts relative to the ear anatomy. This information may be useful for different relevant tasks when programming or fitting CI devices, i.e. it may be useful to deactivate electrodes \cite{Zhao:2016} or to reduce their current strength in electric-acoustic stimulation (EAS) subjects \cite{Imsiecke:2018,Krueger2017}. 
CI arrays use contacts spacing ranging from around \unit[0.8]{mm} to \unit[2]{mm}. CBCT scanners typically used in the clinical routine have a resolution ranging from 0.1 to 0.3 $\text{mm}^{3}$. This relatively low resolution in relation to the contact spacing together with image reconstruction distortions and artifacts in the surrounding of the electrode array cause that manual localization of individual electrode contacts from CBCT scans becomes inaccurate. This manual labor is tedious and time consuming  and not applicable for large data-sets \cite{HacFal2013a}.
For this purpose an automatic method to assess electrode location (e.g. \cite{Bennink2017AutomaticLO,Braithwaite:2016,Noble2015,Zhao:2018a}) would be very useful, but remains challenging. 
The main contributions in this paper are a synthetic CI data-set generator, the MRF-based CI localization algorithm (MRF-A) and its comparison with state-of-the-art algorithms.

\section{State-of-the-Art}
Two categories of data-sets brought up different strategies to automatically locate electrodes from CBCT volumes. The first one contains closely-spaced EAs, i.e. CBCT scans in which there is no contrast difference between electrode contacts, while vice versa there are distantly-spaced EAs with varying intensities between electrode contacts. Noble et al. \cite{Noble:2011} apply gradient vector flow snakes to locate 3D centerlines (as in \cite{Hachmann:2018}), on which the EA is registered by a thin-plate spline transformation. Other algorithms use a blob detector as Frangi et al. \cite{Frangi:1998} or a surface-based centerline point extraction similar to Bouix et al. \cite{Bouix:2005} as a preprocessing step. For instance, Noble et al. \cite{Noble2015} propose a graph-based algorithm that sequentially adds centerline candidates to a growing set of paths, which are than evaluated and pruned to a fixed number of candiates. The predicted EA is refined by a second run using new candidates sampled from a grid around the previously found EA locations. 
Braithwaite et al. \cite{Braithwaite:2016} focused on the problem of locating distantly-spaced electrode arrays, by individually locating electrodes in a given volume of interest (VOI). Three filters are sequentially applied: a thresholding filter, a spherical filter and a Gaussian filter. Detected contacts are reordered based on a prior knowledge of the type of electrode array. Bennink et al. \cite{Bennink2017AutomaticLO} use a filter which enhances small blob-like structures (Lindeberg \cite{Lindeberg:1994}), followed by a curve tracking stage that sequentially includes maximum intensity voxels form a predicted VOI. The estimated curve is smoothed and correlation with the CI specifications is conducted to estimate the final electrode location. More recently, Chi et al. \cite{Chi2019} used conditional generative adversarial networks (cGANs) to generate likelihood maps in which voxel values are proportional to the distance to the nearest candidate contact. Using a high threshold on the map, electrode positions can be extracted. With decreasing threshold, electrodes are merged together, which is tracked and the connection of electrodes is found. 
All preexisting algorithms localize electrodes sequentially, while our approach is the first to jointly optimize all electrode positions, mutual distances and angles.
\section{Synthetic and real CBCT Data-Sets}
Post-operative temporal bone CBCT scans were collected from ten CI users denoted as CBCT1 to CBCT10. The CI users were implanted with a MidScala (CBCT1 and CBCT2) or a SlimJ (CBCT3 to CBCT10) electrode array (Advanced  Bionics,  Valencia,  CA). The arrays include 16 electrode contacts that are equally spaced with a distance of approximately \unit[1]{mm}. CBCT scans were collected using a Xoran MiniCat (Ann Arbor,  MI, USA) for CBCT1 and CBCT2 and a 3D Accuitomo 170 (J. Morita. Mfg. Corp., Kyoto, Japan) for CBCT3 to CBCT10 with an isotropic voxel resolution of 
\unit[0.3]{mm} and \unit[0.125]{mm}, respectively. The manually determination of electrode localization was performed by a clinical expert using the OsiriX MD software (Pixmeo, Geneva, Switzerland) for 3D reconstruction and visualization. 

Similar to Bouix et al. \cite{Bouix:2005} we evaluate algorithms on synthetic data-sets with the advantage that the ground truth (GT) is precisely known. An EA is characterized as a sequence of electrode contacts, each causing high intensity values in CBCT data-sets. Wire leads, receiver coils, and bone structures can have similar appearance, especially in low intensity \textit{lCTs} \cite{Zhao:2019}. A EA follows a helix structure, due to the anatomy of the cochlea. We create synthetic EAs in a two-step procedure: (I) GT electrode positions $x_n$ are generated, which (II) are used to create a synthetic CBCT volume $I_{syn}$.

(I) Assume a Cartesian coordinate system with axis $u,v,w$. Starting at an initial seed position $x_{n=0}$, the EA is created by successively adding further electrodes with every iteration $n$. In
\begin{multline}
  x_{n+1}(u,v,\lambda_1\cdot(n+1)) = x_{n}(u,v,\lambda_1\cdot n) \\
  + [cos(\lambda_2\cdot n\cdot\alpha),sin(\lambda_2\cdot n\cdot\alpha),1] \cdot l_{prior},
  \label{Eq1}
\end{multline}
$l_{prior}$ is the distance between electrodes and $\lambda_2$-weighted $\alpha$ is an angle controlling increasing curvature with $n$. Thus Eq. (\ref{Eq1}) creates a helix that screws in or out of the $u$,$v$-plane as a factor of $\lambda_1$.
These electrode positions are altered by several parameters: mirroring ($\lambda_{mirror} \in \{0,1\}$) to imitate left or right ear CIs and 3D rotations ($\theta_u$,$\theta_v$,$\theta_w$) for arbitrary orientations. These synthetic EA positions are considered GT.

 (II) We create an empty equilateral high resolution synthetic volume $I_{0}$ that comprises all EA contacts positions, also adding a guard interval at all borders. 20 electrode positions $x_n$ and a number of $n_{bones}$ randomly inserted bone-like structures $x_b$ are added by
\begin{equation}  
    I_{0}(x \in \{ x_n \cup x_b \}) = 1 \land I_{0}(x \notin \{ x_n \cup x_b \}) = 0.  
\end{equation} 
$I_{0}$ is smoothed by a Gaussian filter $G_{\sigma_1}(\cdot)$ with standard deviation $\sigma_1$, downsampled by $S_{\downarrow}(\cdot)$ to a target size using cubic interpolation. Intensities are scaled and shifted by $\lambda_3$ and $\lambda_4$ and white noise $N_1$ is added to generate
$I_{syn} = \lambda_3 \cdot S_{\downarrow}(G_{\sigma_1}(I_{0}))+ \lambda_4 + N_1.$
A diverse CI data-set is generated ([$I_{syn}$, GT]-pairs, see Fig \ref{CT_datasets} and Table \ref{L2_table}) using randomly sampled parameters ($\lambda_{1-4}$, $\lambda_{mirror}$, $\theta_{u-w}$, $\alpha_{n}$, $\sigma_1$, $n_{bones}$ and $N_1$) from suitable ranges. The synthetic data-set and its generator are available\footnote{\url{https://github.com/hendrik-hachmann/synCIg}}.
\section{Electrode Array Localization}
We propose a MRF-based algorithm (MRF-A), that estimates the 3D positions of electrodes inside a user specified VOI with the most basal contact marked. The VOI size is around \unit[1]{$\text{cm}^3$} and must contain all EA contacts. The MRF model energy is defined as
\begin{equation}
  E(x) = \sum\nolimits_{s\in \mathcal{V}} \psi_{s}(x_s) + 
  \sum\nolimits_{s\in\mathcal{V}}\sum\nolimits_{t\in \mathcal{N}_{s}} \psi_{s,t}(x_s,x_t),
  \label{eq:model}
\end{equation}
in which $\mathcal{V}$ is a set of nodes with neighbors $\mathcal{N}_s$. The random variable $x_s$ is the 3D position of node $s$.
The unary potential
\begin{equation}
  \psi_{s}(x_s) = \Theta_{1} \cdot G_{\sigma_2}(I({x}_{s})) + \Theta_{2} \cdot G_{Blob}(I({x}_{s}))
\end{equation}
takes a CBCT intensities $I$ into account, where $G_{\sigma_2}(I)$ is a smoothed (Gaussian filter, standard deviation $\sigma_2$) and $G_{Blob}(I)$ a blob filter-enhanced version of $I$. All $\Theta_{x}$ are empirically set weighting parameters.
The binary potential
\begin{dmath*}
  \psi_{s,t}(x_s,x_t) = 
  \left\lbrace
      \begin{array}{ll}
         \Theta_{3}\left\|{x}_{s} - {x}_{t}-d_{st1}\right\|_{2}^{2}\condition[]{for $t \in \mathcal{N}_{1,s}$}
         \\[2mm]
         \Theta_{4}\left\|{x}_{s} - {x}_{t}-d_{st2}\right\|_{2}^{2}\condition[]{for $t \in \mathcal{N}_{2,s}$}
      \end{array}
  \right.
\end{dmath*}
models relationships between node ${x}_{s}$ and a neighboring node ${x}_{t}$ in the EA. The neighborhood $\mathcal{N}_{1,s}$ is used to constrain the node distance to match the prior known distance $d_{st1}$ for a specific EA. 
\begin{figure}[t]
  \centering
  \includegraphics[height=0.13\textwidth]{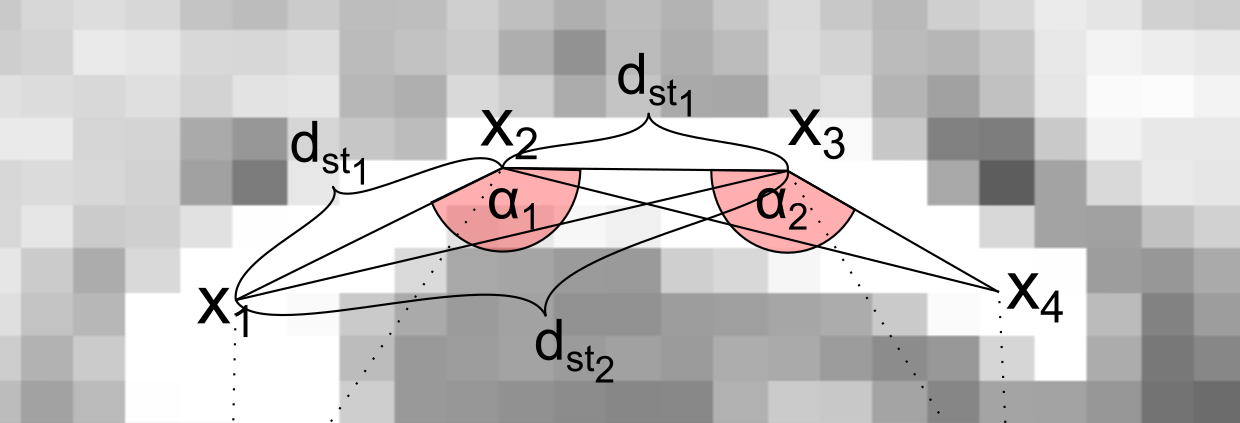} 
  \caption{Binary potential in MRF-A: The distance ratio $d_{st_2}/d_{st_1}$ between nodes $x_n$ constrains the angles $\alpha_n$.}
 \label{angularTerm}
\end{figure}
The second order neighborhood $\mathcal{N}_{2,s}$ is used to form an angular constraint as illustrated in Fig. \ref{angularTerm}. We enforce an decreasing angle $\alpha_n$ of the EA in apical direction by successively reducing the ratio $d_{st_2}/d_{st_1}$. 

In order to efficiently minimize the MRF energy (Eq. (\ref{eq:model})), we approximate the maximum a posterior (MAP) probability using of max-product particle belief propagation (PBP) as Pacheco et al. \cite{Pacheco:2011}. Following the notation of Besse et al. \cite{besse2012pmbp}, for a set of particles $(i)$ and PBP iteration $n \in 1,\cdots,N$ we minimize the log disbelief $B_{s}^{n}(x_{s}^{(i)})$ to obtain the most likely configuration $\hat{x}_{s} = \argmin B_{s}^{N}(x_{s}^{(i)})$ after $N$ iterations of belief propagation. Since our binary potential $\psi_{s,t}$ consists of loops (e.g. $\overline{x_{1}x_{2}}$, $\overline{x_{2}x_{3}}$, and $\overline{x_{1}x_{3}}$ in Fig. \ref{angularTerm}) we optimize via message passing (Wainwright et al. \cite{Wainwright:2005}):
\begin{equation*}
  \begin{split}
    B_{s}^{n}(x_{s}^{(i)}) &= \psi_{s}(x_{s}^{(i)}) + \sum\nolimits_{t\in \mathcal{N}_s}M_{t \rightarrow s}^{n}(x_s^{(i)}) \text{ and}\\
    M_{t \rightarrow s}^{n}(x_s^{(i)}) &= \underset{x_{t} \in P_t}\min[\psi_{s,t}(x_s,x_t)+B_t^{n-1}(x_t)-M_{s \rightarrow t}^{n-1}(x_t)].
    \label{eq:lengths}
 \end{split} 
\end{equation*}
$B_{s}^{n}(x_{s}^{(i)})$ is iteratively calculated by messages $M_{t \rightarrow s}^{n}(x_s^{(i)})$ sent from node $t$ to node $s$ at iteration $n$. 

The performance of PBP is heavily dependent on sampling \textit{good} particles $x^n_s$ from the target log-disbelief distribution $x^n_s \sim B_s^n(x_s)$. Each particle $x^n_s$ is sampled using Markov chain Monte Carlo (MCMC) sampling $\{x_s^{(i)<m>}\}_{m=1,\cdots,M}$ with $m$ MCMC iterations. We apply slice sampling particle belief propagation (S-PBP) as in M\"{u}ller et al. \cite{MueYan2013}. Slice sampling estimates an interval $A$ on the estimated belief distribution, which defines a slice $u$. The interval $A$ can be computed analytically from the potentials $\psi_s$ and $\psi_{s,t}$, and particles $x^n_s$ are sampled from the uniform distribution $\mathcal{U}_{A^{<m>}}$. 

However, PBP gives the opportunity to encode further CI a priori knowledge in the form of particle sets that guides the optimization and improves convergence. In each PBP iteration $n$, the most likely configuration $\hat{x}_{t}$ is used to augment the current slice sampling particle set $P_t = \{x_t^{(1)},\cdots,x_t^{(p)}\}$ of node $t$ containing $p$ particles: 
$P_{t,augmented} = P_t \cup P_{mobility} \cup P_{CI} \cup P_{rotated} \cup P_{knn}$.
At each PBP iteration, $P_{mobility}$ adds the positions $\hat{x}_{t}$ of each direct neighbor to the particle set as well as predicted positions at the apical end of the EA, based on position, length and angle of the two most apical nodes. In that way, $P_{mobility}$ provides additional mobility to the MRF model in longitudinal direction along the EA. The particles sets $P_{CI}$ attempt to predict the whole EA making use of its helix shape. From the positions $\hat{x}_{t}$, a low energy section (evaluating Eq. (\ref{eq:model})) of 4 to 6 consecutive nodes is extracted, angles in this sequence are calculated and a helix shape is fit. The predicted helix positions are added to $P_{CI}$. Electrode positions may be predicted in the \textit{wrong way direction}, e.g. predicting a clockwise CI counter-clockwise. To encounter this problem, we add $P_{rotated}$ to the particle set, a 180 degree rotated version of $\hat{x}_{t}$ along the vector: basal node $x_{0}$ to center position of $\hat{x}_{t}$. The particle set $P_{knn}$ includes the $k$ nearest candidate points, generated by the algorithm of Broix et al. \cite{Bouix:2005}, and thus attracts model towards these locations. The particle set is decimated by diverse particle max-product \cite{pacheco15} to remain constant in size.
\begin{figure}[t]
  \includegraphics[height=0.235\textheight,bb=55 17 534 583, clip=true]{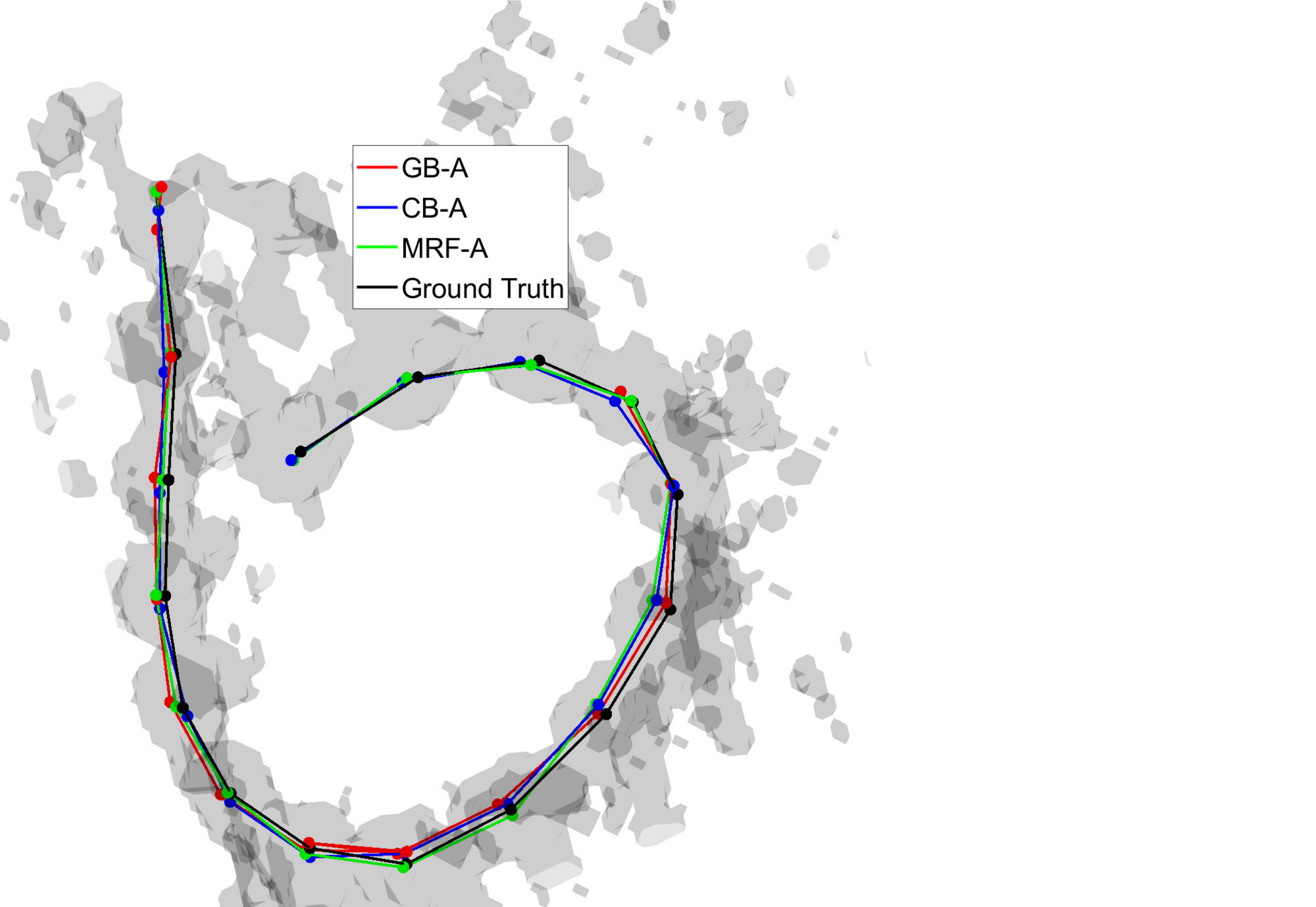}
  \includegraphics[height=0.235\textheight,bb=1 1 400 565, clip=true]{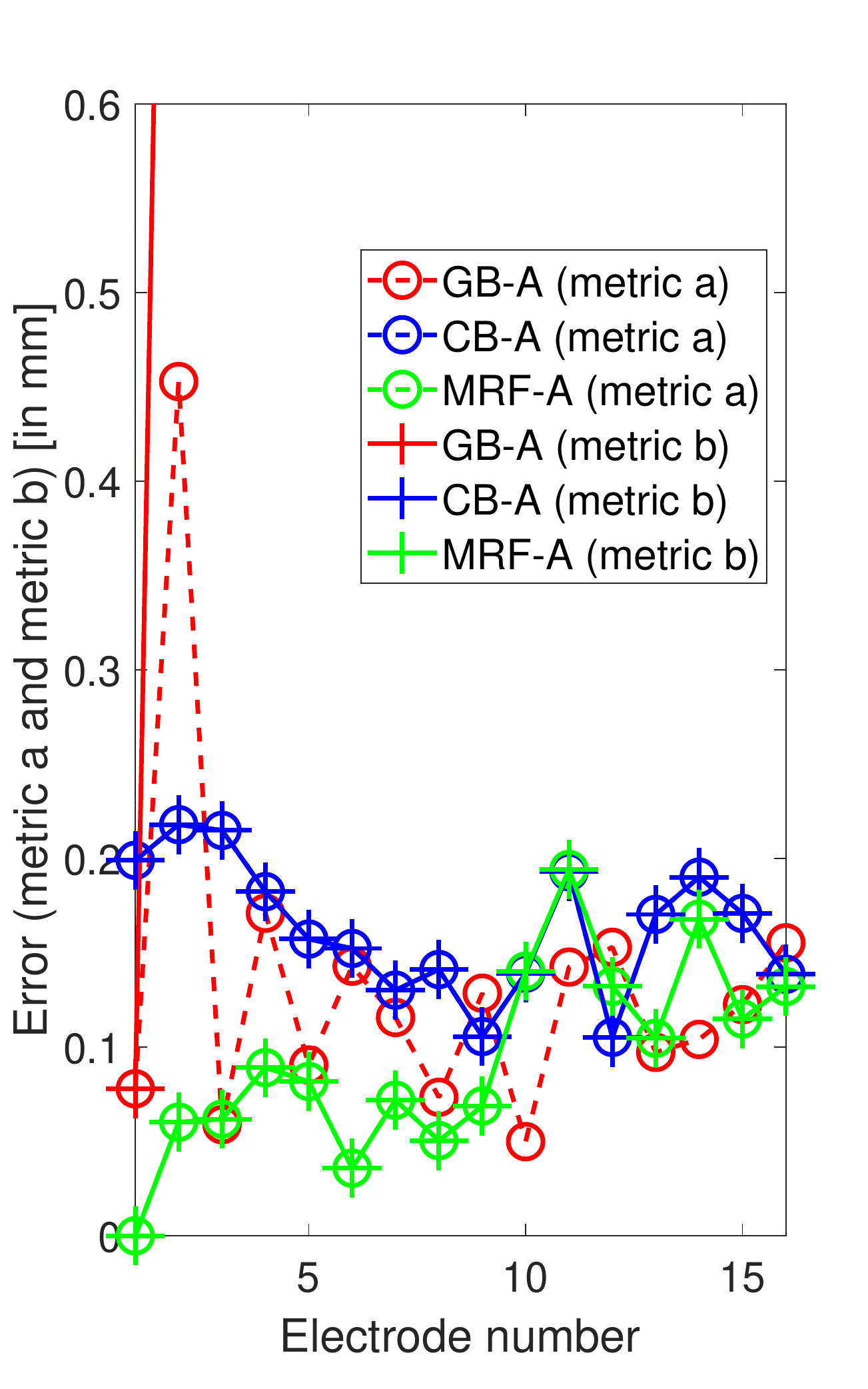}
  \caption{Qualitative comparison of the algorithms GB-A (based on \cite{Noble2015}), CB-A (based on \cite{Bennink2017AutomaticLO}), MRF-A and the ground truth (GT) on the data-set \textit{CBCT3}. The corresponding per electrode localization errors along the electrode array (EA) are given on the right. Index 1 is the most basal electrode. For CB-A and MRF-A dashed and solid lines run coincidently. In Table \ref{L2_table} metric a) and b) errors are averages per data-set.}
  \label{3d_results}
\end{figure}
Due to its stochastic nature, each optimization of the particle-based MRF can be seen as a random walk of the CI model, thus providing different estimates at multiple runs. This can be used in a random search, e.g. run the algorithm $N_{runs}$ times and take the best solution in terms of lowest MAP energy. Each MRF run is initialized at the most basal electrode and pointing randomly inwards the VOI. The current MRF state $\hat{x}_{t}$ can degenerate, e.g. reach a folded prediction of the EA, and no change in MAP energy is detected. In this case we reinitialize ties between electrode positions by calculating the shortest path between electrodes. 

\section{Experiments and Results}
\label{experiments}
\newcommand{\myVal}{6.3mm}
\newcommand{\myVall}{6.5mm}
\begin{sidewaystable}
  \centering
 \begin{tabular}{
  |p{12mm}
  |p{\myVall}
  |p{\myVal}
  |p{\myVal}
  |p{\myVall}
  |p{\myVal}
  |p{\myVal}
  |p{\myVall}
  |p{\myVal}
  |p{\myVal}
  ||p{12mm}
  |p{\myVall}
  |p{\myVal}
  |p{\myVal}
  |p{\myVall}
  |p{\myVal}
  |p{\myVal}
  |p{\myVall}
  |p{\myVal}
  |p{\myVal}|
  }
   Data-set & \multicolumn{3}{c|}{GB-A}  & \multicolumn{3}{c|}{CB-A} & \multicolumn{3}{c||}{MRF-A} & Data-set & \multicolumn{3}{c|}{GB-A}  & \multicolumn{3}{c|}{CB-A}  & \multicolumn{3}{c|}{MRF-A}\\   
  \hline  
  & \textbf{mean} & \hspace{2mm}a) & \hspace{2mm}b) & \textbf{score} & \hspace{2mm}a) & \hspace{2mm}b) & \textbf{score} & \hspace{2mm}a) & \hspace{2mm}b) & & \textbf{score} & \hspace{2mm}a) & \hspace{2mm}b) & \textbf{score} & \hspace{2mm}a) & \hspace{2mm}b) & \textbf{score} & \hspace{2mm}a) & \hspace{2mm}b)\\
  \hline
  \hline
\textit{CBCT1} & 1.19 & 0.33 & 2.06 &0.57 & 0.41 & 0.73 &\textbf{0.36} & 0.36 & 0.36 & \textit{Synth1} &0.65 & 0.39 & 0.92	&0.90 & 0.39 & 1.41 & \textbf{0.25} & 0.25 & 0.25\\
\textit{CBCT2} & \textbf{0.41}& 0.40 &0.41 & 1.32 & 0.63 & 2.00 & 1.24 & 0.49 & 1.99 & \textit{Synth2} & 0.43 & 0.43 & 0.43	& 0.73 & 0.58 & 0.89 & \textbf{0.24} & 0.24 & 0.24\\
\textit{CBCT3} & 1.15 & 0.13 & 2.16 & 0.16 & 0.16 & 0.16 & \textbf{0.09} & 0.09 & 0.09 & \textit{Synth6} & 0.55 & 0.39 & 0.72 & 1.92 & 0.51 & 3.32 & \textbf{0.24} & 0.24 & 0.24\\
\textit{CBCT4} & \textbf{0.24} & 0.24 & 0.24 & 0.27 & 0.27 & 0.27 & \textbf{0.24} & 0.24 & 0.24 & \textit{Synth8} & 0.58 & 0.52 & 0.64 & 0.30 & 0.30 & 0.30 & \textbf{0.23} & 0.23 & 0.23\\
\textit{CBCT5} & \textbf{0.34} & 0.34 & 0.34 & 0.63 & 0.44 & 0.83 & 0.46 & 0.36 & 0.56 & \textit{Synth10} & 0.62 & 0.54 & 0.70 & \textbf{0.26} & 0.26 & 0.26 & 0.51  & 0.43 & 0.59\\
\textit{CBCT6} & \textbf{0.34} & 0.16 & 0.52 & 1.27 & 0.21 & 2.34 & 1.41 & 0.22 & 2.59 & \textit{Synth11} & 0.91 & 0.22 & 1.61 & 0.96 & 0.41 & 1.52 & \textbf{0.82} & 0.80 & 0.83\\
\textit{CBCT7} & 1.06 & 0.28 & 1.85 & 0.27 & 0.27 & 0.27 & \textbf{0.26} & 0.26 & 0.26 & \cellcolor{yellow!25}\textit{Synth3} & \cellcolor{yellow!25}0.93 & \cellcolor{yellow!25}0.25 & \cellcolor{yellow!25}1.61 & \cellcolor{yellow!25}0.45 & \cellcolor{yellow!25}0.37 & \cellcolor{yellow!25}0.52 & \cellcolor{yellow!25}\textbf{0.29} & \cellcolor{yellow!25}0.29 & \cellcolor{yellow!25}0.29\\
\textit{CBCT8} & 0.32 & 0.32 & 0.32 & \textbf{0.31} & 0.31 & 0.32 & 0.51 & 0.37 & 0.65 & \cellcolor{yellow!25}\textit{Synth7} & \cellcolor{yellow!25}0.58 & \cellcolor{yellow!25}0.29 & \cellcolor{yellow!25}0.87 &\cellcolor{yellow!25}\textbf{0.25} & \cellcolor{yellow!25}0.25 & \cellcolor{yellow!25}0.25 & \cellcolor{yellow!25}\textbf{0.25} & \cellcolor{yellow!25}0.25 & \cellcolor{yellow!25}0.25\\
\textit{CBCT9} & 0.70 & 0.20 & 1.20 & 1.48 & 0.42 & 2.54 & \textbf{0.17} & 0.17 & 0.17 & \cellcolor{yellow!25}\textit{Synth9} & \cellcolor{yellow!25}0.52 & \cellcolor{yellow!25}0.41 & \cellcolor{yellow!25}0.64 & \cellcolor{yellow!25}0.67 & \cellcolor{yellow!25}0.30 & \cellcolor{yellow!25}1.04 & \cellcolor{yellow!25}\textbf{0.43} & \cellcolor{yellow!25}0.40 & \cellcolor{yellow!25}0.45\\
\textit{CBCT10} & 1.52 & 0.28 & 2.76 & 0.69 & 0.31 & 1.07 & \textbf{0.25} & 0.25 & 0.25) & \cellcolor{yellow!50}\textit{Synth4} & \cellcolor{yellow!50}0.39 & \cellcolor{yellow!50}0.35 & \cellcolor{yellow!50}0.43 & \cellcolor{yellow!50}1.10 & \cellcolor{yellow!50}0.35 & \cellcolor{yellow!50}1.84 & \cellcolor{yellow!50}\textbf{0.28} & \cellcolor{yellow!50}0.28 & \cellcolor{yellow!50}0.28\\
& & & & & & & & &  & \cellcolor{yellow!50}\textit{Synth5} & \cellcolor{yellow!50}0.58 & \cellcolor{yellow!50}0.31 & \cellcolor{yellow!50}0.86 & \cellcolor{yellow!50}0.69 & \cellcolor{yellow!50}0.38 & \cellcolor{yellow!50}0.99 & \cellcolor{yellow!50}\textbf{0.24} & \cellcolor{yellow!50}0.24 & \cellcolor{yellow!50}0.24\\
\hline 
\textbf{Mean} & 0.73 & 0.27 & 1.19 & 0.70 & 0.34 & 1.05 & \textbf{0.50} & 0.28 & 0.72 & \textbf{Mean} & 0.61 & 0.37 & 0.86 & 0.75 & 0.37 & 1.12 & \textbf{0.34} & 0.33 & 0.36\\%
\textbf{Median} & 0.55 & 0.28 & 0.86 & 0.60 & 0.31 & 0.78 & \textbf{0.31} & 0.25 & 0.31 & \textbf{Median} & 0.58 & 0.39 & 0.72 & 0.69 & 0.37 & 0.99 & \textbf{0.25} & 0.25 & 0.25\\%
  \end{tabular}
  \caption{EA localization score, metric a) and metric b) [in mm] measured for 11 synthetic and 10 CBCT data-sets. On CBCT data-sets, the MRF-A achieves a mean score of \unit[0.50]{mm}, which is 28.6\% better than the second best algorithm (median: \unit[0.31]{mm}, 43.6\% increase; 31.5\% and 48.6\% compared to third method). On synthetic data-sets MRF-A achieves a mean score of 0.34, a 44.3\% to 54.7\% mean and a 56.9\% to 63.8\% median increase. While most EAs can be classified as distantly spaced, the yellow highlighted data-sets have increasingly closer spaced EAs. No data-set has a pure no contrast EA. Average runtimes for MRF-A are 21 minutes ($N_{runs}$ set to 100), GB-A 2 minutes ($\alpha_9 = 3$ and $P = 500$) and CB-A 1.4 seconds, using \MATLAB implementations and a Xeon W-2145 processor.
  }
  \label{L2_table}
\end{sidewaystable}
 MRF-A is compared with a graph-based algorithm GB-A, similar to Noble et al. \cite{Noble2015} and with a correlation-based algorithm CB-A, similar to Bennink et al. \cite{Bennink2017AutomaticLO}. Parameters of all algorithms are optimized empirically, for GB-A and CB-A starting in the vicinity of the parameters specified in the corresponding papers. We perform the algorithm evaluation based on two localization error metrics. For each predicted electrode, we calculate the Euclidean distance a) to the nearest GT electrode position and b) to the GT electrode position with the same label or electrode number. While metric a) represents a mean electrode localization error, it does not appropriately reflect errors like EA folding, since the folded electrodes can be near other GT positions. Also if the whole predicted EA is shifted by one electrode, this error is underestimated. Thus, we evaluate on metric b) as well. Using the average of metric a) and b) we create an overall score. Results can be seen in Table \ref{L2_table}. 
 Although MRF-A needs a lot of computation power, the EA localization is an automatic task where runtimes are of minor importance. Fig. \ref{3d_results} presents a comparison of the localization results obtained with the GB-A, CB-A and MRF-A on the \textit{CBCT3} data-set, with corresponding errors on metrics a) and b) along the EA shown on the right. In this example, all three algorithms provide a low metric a) scores, while GB-A has two electrode detections at the most basal electrode, leading to further electrodes being shifted from the GT and larger than \unit[0.6]{mm} metric b) errors. Similar errors can be found for all algorithms in Table \ref{L2_table}, indicated by high metric b) values.

\section{Conclusion}
In this work, we have proposed an MRF-based algorithm (MRF-A) for CI electrode localization that optimizes the EA contacts positions by satisfying the CI constraints: high CBCT intensity and blob-shape of electrodes, distance and angle between electrodes as well as angle increase in apical direction. The particle-based optimization is guided by several CI-adapted heuristics to increase convergence. Using the stochastic nature of MRF-A, we created a pool of candidate EAs and the best candidate is selected. 
We compared our algorithms with two implementations based on state-of-the-art algorithms (GB-A, CB-A) using synthetic data-sets, for which a precise GT is known, and real CBCT data-sets from 10 subjects. Two metrics show that MRF-A is robust and achieves low error rates.  

\section{Compliance with ethical standards}
This is a numerical simulation study for which no ethical approval was required.





\bibliographystyle{IEEEbib}

\bibliography{mybibliography}

\end{document}